\documentclass{article}
\usepackage{}
\usepackage{}
\usepackage{amsmath, algorithm, graphicx, subcaption, siunitx, float, hyperref, cleveref, amssymb}
\usepackage[noend]{algpseudocode}
\usepackage[utf8]{inputenc}

\makeatletter
\def\BState{\State\hskip-\ALG@thistlm}
\makeatother

\hypersetup{
    colorlinks=true,
    linkcolor=black,
    citecolor=black,
    urlcolor=blue,
}

\usepackage{natbib}
\setcitestyle{authoryear,open={(},close={)}}

\newcommand{\changeurlcolor}[1]{\hypersetup{urlcolor=#1}} 

\nocite{*}

\title{{\Large Phasic Policy Gradient}}

\author{
  \small \textbf{\scriptsize Karl Cobbe}\\
  \small \texttt{\scriptsize karl@openai.com}
  \and
  \small \textbf{\scriptsize Jacob Hilton}\\
  \small \texttt{\scriptsize jhilton@openai.com}
  \and
  \small \textbf{\scriptsize Oleg Klimov}\\
  \small \texttt{\scriptsize oleg@openai.com}
  \and
  \small \textbf{\scriptsize John Schulman}\\
  \small \texttt{\scriptsize joschu@openai.com}
}
\date{}

\begin{document}

\maketitle

\begin{abstract}

We introduce Phasic Policy Gradient (PPG), a reinforcement learning framework which modifies traditional on-policy actor-critic methods by separating policy and value function training into distinct phases. In prior methods, one must choose between using a shared network or separate networks to represent the policy and value function. Using separate networks avoids interference between objectives, while using a shared network allows useful features to be shared. PPG is able to achieve the best of both worlds by splitting optimization into two phases, one that advances training and one that distills features. PPG also enables the value function to be more aggressively optimized with a higher level of sample reuse. Compared to PPO, we find that PPG significantly improves sample efficiency on the challenging Procgen Benchmark.

\end{abstract}

\section{Introduction}

Model free reinforcement learning (RL) has enjoyed remarkable success in recent years, achieving impressive results in diverse domains including DoTA \citep{dota}, Starcraft II \citep{starcraft}, and robotic control \citep{rubiks}. Although policy gradient methods like PPO \citep{ppo}, A3C \citep{mnih2016asynchronous}, and IMPALA \citep{impala} are behind some of the most high profile results, many related algorithms have proposed a variety of policy objectives \citep{trpo, acktr, awr, vmpo, ddpg, sac}. All of these algorithms fundamentally rely on the actor-critic framework, with two key quantities driving learning: the policy and the value function. In practice, whether or not to share parameters between the policy and the value function networks is an important implementation decision. There is a clear advantage to sharing parameters: features trained by each objective can be used to better optimize the other.

However, there are also disadvantages to sharing network parameters. First, it is not clear how to appropriately balance the competing objectives of the policy and the value function. Any method that jointly optimizes these two objectives with the same network must assign a relative weight to each. Regardless of how well this hyperparameter is chosen, there is a risk that the optimization of one objective will interfere with the optimization of the other. Second, the use of a shared network all but requires the policy and value function objectives to be trained with the same data, and consequently the same level of sample reuse. This is an artificial and undesirable restriction.

We address these problems with Phasic Policy Gradient (PPG), an algorithm which preserves the feature sharing between the policy and value function, while otherwise decoupling their training. PPG operates in two alternating phases: the first phase trains the policy, and the second phase distills useful features from the value function. More generally, PPG can be used to perform any auxiliary optimization alongside RL, though in this work we take value function error to be the sole auxiliary objective. Using PPG, we highlight two important observations about on-policy actor-critic methods:

\begin{enumerate}
  \item Interference between policy and value function optimization can negatively impact performance when parameters are shared between the policy and the value function networks.
  \item Value function optimization often tolerates a significantly higher level of sample reuse than policy optimization.
\end{enumerate}

By mitigating the interference between the policy and value function objectives while still sharing representations, and by optimizing each with the appropriate level of sample reuse, PPG significantly improves sample efficiency.

\section{Algorithm} \label{sec:algorithm}

In PPG, training proceeds in two alternating phases: the \textit{policy phase}, followed by the \textit{auxiliary phase}. During the policy phase, we train the agent with Proximal Policy Optimization (PPO) \citep{ppo}. During the auxiliary phase, we distill features from the value function into the policy network, to improve training in future policy phases. Compared to PPO, the novel contribution of PPG is the inclusion of periodic auxiliary phases. We now describe each phase in more detail.

\begin{figure*}
\centering
\includegraphics[width=.4\textwidth]{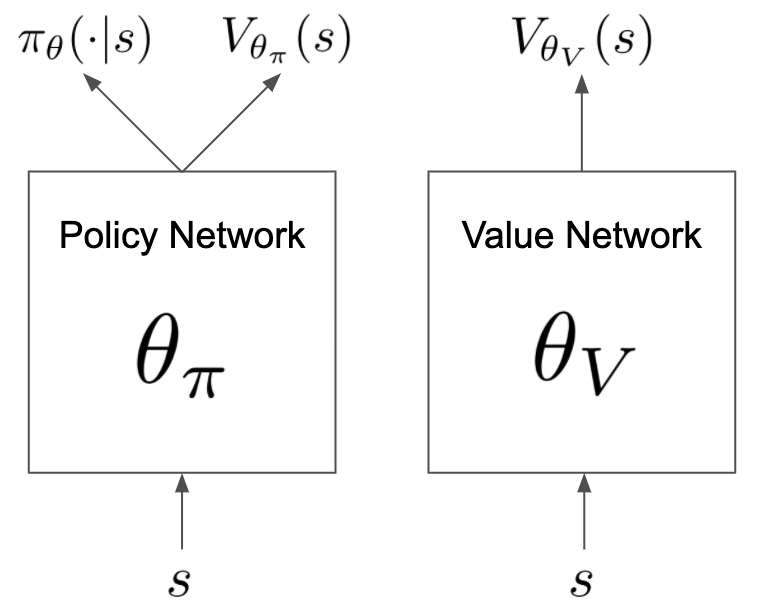}
\caption{PPG uses disjoint policy and value networks to reduce interference between objectives. The policy network includes an auxiliary value head.}
\label{fig:ppg_arch}
\end{figure*}

During the policy phase, we optimize the same objectives from PPO, notably using disjoint networks to represent the policy and the value function (\Cref{fig:ppg_arch}). Specifically, we train the policy network using the clipped surrogate objective $$L^{clip} = \mathbb{\hat{E}}_t\left[\min(r_t(\theta)\hat{A_t}, \text{clip}(r_t(\theta), 1 - \epsilon, 1 + \epsilon)\hat{A_t})\right]$$ where $r_t(\theta) = \frac{\pi_\theta(a_t | s_t)}{\pi_{\theta_{old}}(a_t | s_t)}$, and $\hat{A_t}$ is an estimator of the advantage function at timestep $t$. We optimize $L^{clip} + \beta_SS[\pi]$, where $\beta_S$ is a constant and $S$ is a an entropy bonus for the policy. To train the value function network, we optimize

$$L^{value} = \mathbb{\hat{E}}_t\left[\frac{1}{2}(V_{\theta_V}(s_t) -  \hat{V}^{\text{targ}}_{t})^2\right]$$
where $\hat{V}^{\text{targ}}$ are value function targets. Both $\hat{A}$ and $\hat{V}^{\text{targ}}$ are computed with GAE \citep{gae}.

During the auxiliary phase, we optimize the policy network with a joint objective that includes an arbitrary auxiliary loss and a behavioral cloning loss:

$$L^{joint} = L^{aux} + \beta_{clone} \cdot \mathbb{\hat{E}}_t\left[KL[\pi_{\theta_{old}}(\cdot|s_t),\pi_{\theta}(\cdot|s_t)]\right]$$
where $\pi_{\theta_{old}}$ is the policy right before the auxiliary phase begins. That is, we optimize the auxiliary objective while otherwise preserving the original policy, with the hyperparameter $\beta_{clone}$ controlling this trade-off. In principle $L^{aux}$ could be any auxiliary objective. At present, we simply use the value function loss as the auxiliary objective, thereby sharing features between the policy and value function while minimizing distortions to the policy. Specifically, we define $$L^{aux} = \frac{1}{2} \cdot \mathbb{\hat{E}}_t\left[(V_{\theta_{\pi}}(s_t) -  \hat{V}^{\text{targ}}_{t})^2\right]$$ where $V_{\theta_{\pi}}$ is an auxiliary value head of the policy network, shown in \Cref{fig:ppg_arch}.

\begin{algorithm}
\caption{PPG}
\begin{algorithmic}
\For{phase = $1,2,...$}
  \State Initialize empty buffer $B$
  \For{iteration = $1,2,...,N_{\pi}$}\Comment{Policy Phase}
    \State Perform rollouts under current policy $\pi$
    \State Compute value function target $\hat{V}^{\text{targ}}_{t}$ for each state $s_t$
    \For{epoch = $1,2,...,E_{\pi}$}\Comment{Policy Epochs}
        \State Optimize $L^{clip} + \beta_SS[\pi]$ wrt $\theta_\pi$
    \EndFor
    \For{epoch = $1,2,...,E_V$}\Comment{Value Epochs}
        \State Optimize $L^{value}$ wrt $\theta_V$
    \EndFor
    \State Add all ($s_t$, $\hat{V}^{\text{targ}}_{t}$) to $B$
  \EndFor
  \State Compute and store current policy $\pi_{\theta_{old}}(\cdot|s_t)$ for all states $s_t$ in $B$
  \For{epoch = $1,2,...,E_{aux}$}\Comment{Auxiliary Phase}
    \State Optimize $L^{joint}$ wrt $\theta_\pi$, on all data in $B$
    \State Optimize $L^{value}$ wrt $\theta_V$, on all data in $B$
  \EndFor
\EndFor
\end{algorithmic}
\end{algorithm}

This auxiliary value head and policy itself share all parameters except for the final linear layers. The auxiliary value head is used purely to train representations for the policy; it has no other purpose in PPG. Note that the targets $\hat{V}^{\text{targ}}$ are the same targets computed during the policy phase. They remain fixed throughout the auxiliary phase. During the auxiliary phase, we also take the opportunity to perform additional training on the value network by further optimizing $L^{value}$. Note that $L^{value}$ and $L^{joint}$ share no parameter dependencies, so we can optimize these objectives separately. 

We briefly explain the role of each hyperparameter. $N_{\pi}$ controls the number of policy updates performed in each policy phase. $E_{\pi}$ and $E_V$ control the sample reuse for the policy and value function respectively, during the policy phase. Although these are conventionally set to the same value, this is not a strict requirement in PPG. Note that $E_V$ influences the training of the true value function, not the auxiliary value function. $E_{aux}$ controls the sample reuse during the auxiliary phase, representing the number of epochs performed across all data in the replay buffer. It is usually by increasing $E_{aux}$, rather than $E_V$, that we increase sample reuse for value function training. For a detailed discussion on the relationship between $E_{aux}$ and $E_V$, see \Cref{appendix:vf_aux_true}. Default values for all hyperparameters can be found in \Cref{appendix:hyperparameters}. Code for PPG can be found at \href{https://github.com/openai/phasic-policy-gradient}{https://github.com/openai/phasic-policy-gradient}.

\section{Experiments}

\begin{figure*}
\centering
\includegraphics[width=\textwidth]{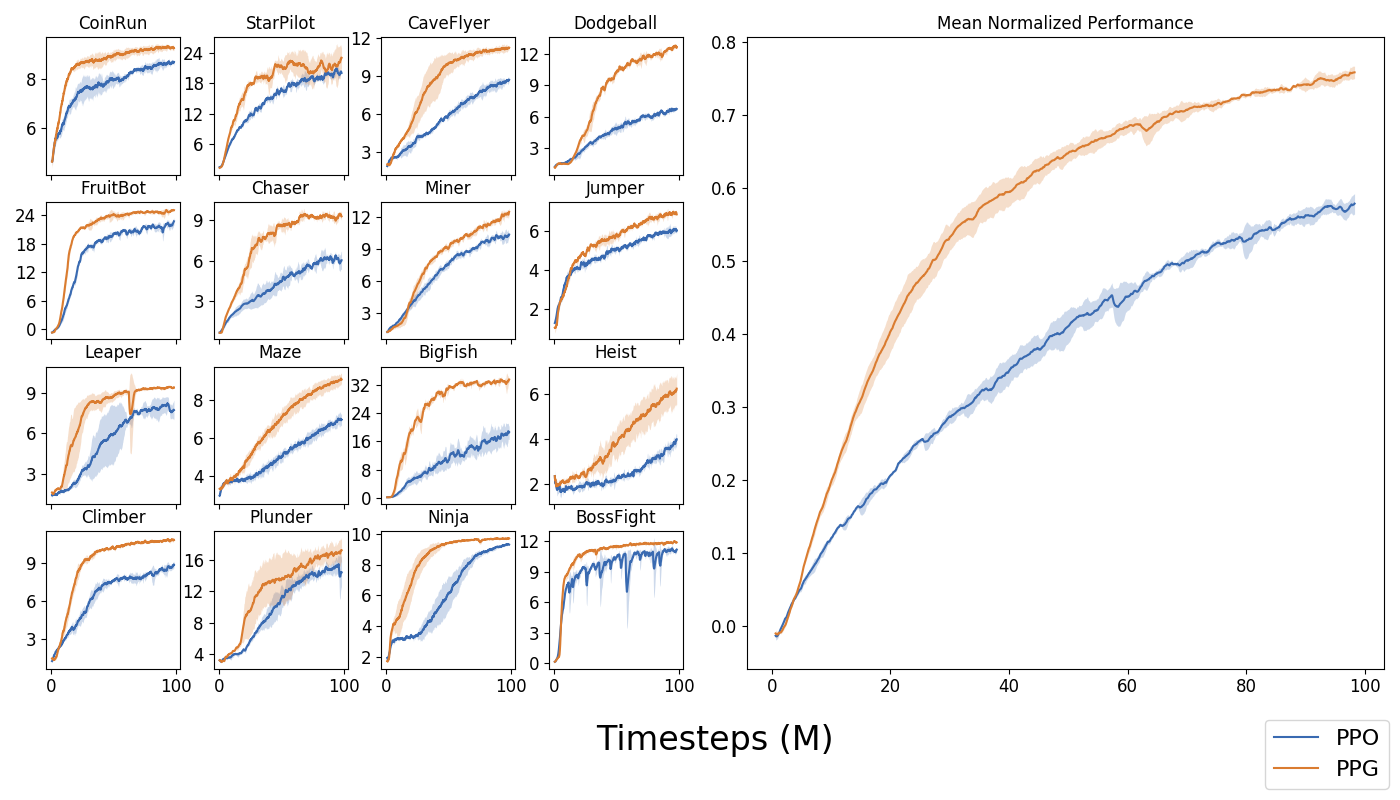}
\caption{Sample efficiency of PPG compared to a PPO baseline}
\label{fig:ppg_vs_ppo}
\end{figure*}

We report results on the environments in Procgen Benchmark \citep{procgen}. This benchmark was designed to be highly diverse, and we expect improvements on this benchmark to transfer well to many other RL environments. Throughout all experiments, we use the hyperparameters found in \Cref{appendix:hyperparameters} unless otherwise specified. When feasible, we compute and visualize the standard deviation across 3 separate runs.

\subsection{Comparison to PPO}

We begin by comparing our implementation of PPG to the highly tuned implementation of PPO from \cite{procgen}. We note that this implementation of PPO uses a near optimal level of sample reuse and a near optimal relative weight for the value and policy losses, as determined by a hyperparameter sweep. Results are shown in \Cref{fig:ppg_vs_ppo}. We can see that PPG achieves significantly better sample efficiency than PPO in nearly every environment.

We have noticed that the importance of representation sharing between the policy and value function does seem to vary between environments. While it is critical to share parameters between the policy and the value function in Procgen environments (see \Cref{appendix:separate_networks}), this is often unnecessary in environments with a lower dimensional input space \citep{sac}. We conjecture that the high dimensional input space in Procgen contributes to the importance of sharing representations between the policy and the value function. We therefore believe it is in environments such as these, particularly those with vision-based observations, that PPG is most likely to outperform PPO and other similar algorithms.

\subsection{Policy Sample Reuse} \label{sec:policy_sr}

\begin{figure*}
\centering
\includegraphics[width=\textwidth]{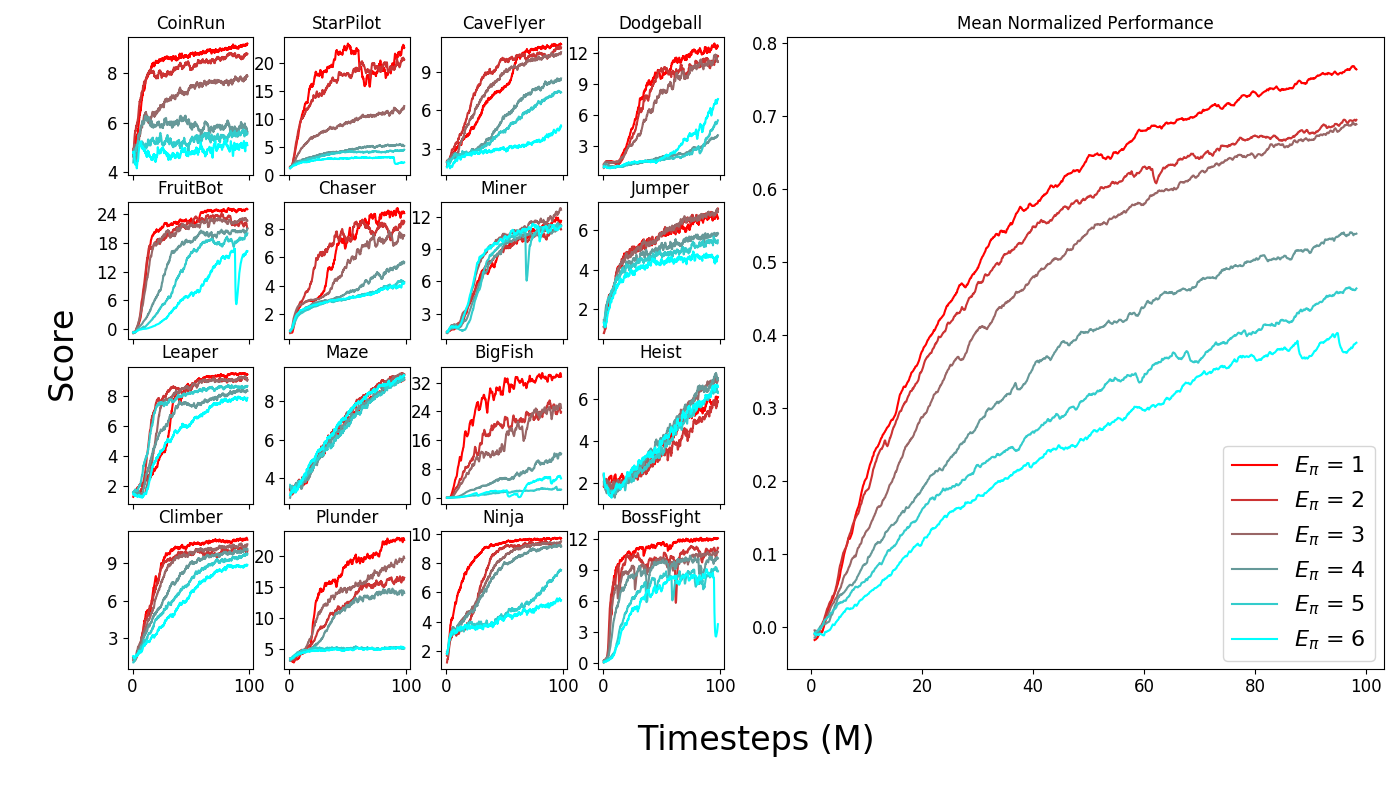}
\caption{Performance with varying levels of policy sample reuse}
\label{fig:pi_sample_reuse}
\end{figure*}

In PPO, choosing the optimal level of sample reuse is not straightforward. Increasing sample reuse in PPO implies performing both additional policy optimization and additional value function optimization. This leads to an undesirable confounding of effects, making it harder to analyze the impact of policy sample reuse alone. Empirically, we find that performing 3 epochs per rollout is best in PPO, given our other hyperparameter settings (see \Cref{appendix:ppo_sample_reuse}).

In PPG, policy and value function training are decoupled, and we can train each with different levels of sample reuse. In order to better understand the impact of policy sample reuse, we choose to vary the number of policy epochs ($E_{\pi}$) without changing the number of value function epochs ($E_V$). Results are shown in \Cref{fig:pi_sample_reuse}.

As we can see, training with a single policy epoch is almost always optimal or near-optimal in PPG. This suggests that the PPO baseline benefits from greater sample reuse only because the extra epochs offer additional value function training. When value function and policy training are properly isolated, we see little benefit from training the policy beyond a single epoch. Of course, various hyperparameters will influence this result. If we use an artificially low learning rate, for instance, it will become advantageous to increase policy sample reuse. Our present conclusion is simply that when using well-tuned hyperparameters, performing a single policy epoch is near-optimal.

\begin{figure*}
\centering
\includegraphics[width=\textwidth]{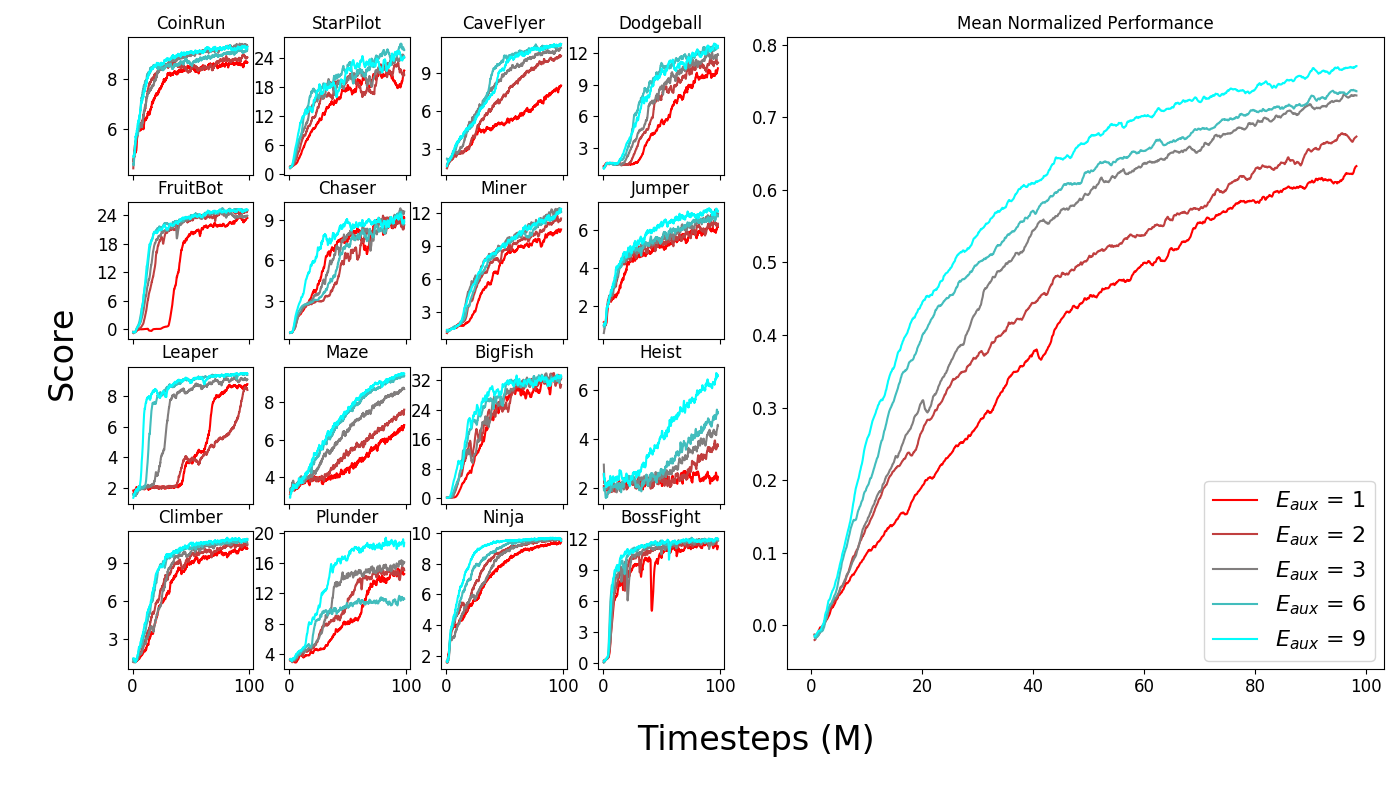}
\caption{Performance with varying levels of value function sample reuse}
\label{fig:nae}
\end{figure*}

\subsection{Value Sample Reuse} \label{sec:value_sr}

We now evaluate how performing additional epochs during the auxiliary phase impacts performance. We expect there to be a trade-off: using too many epochs runs the risk of overfitting to recent data, while using fewer epochs will lead to slower training. We vary the number of auxiliary epochs ($E_{aux}$) from $1$ to $9$ and report results in \Cref{fig:nae}.

We find that training with additional auxiliary epochs is generally beneficial, with performance tapering off around 6 auxiliary epochs. We note that training with additional auxiliary epochs offers two possible benefits. First, due to the optimization of $L^{joint}$, we may expect better-trained features to be shared with the policy. Second, due to the optimization of $L^{value}$, we may expect to train a more accurate value function, thereby reducing the variance of the policy gradient in future policy phases. In general, which benefit is more significant is likely to vary between environments. In Procgen environments, the feature sharing between policy and value networks appears to play the more critical role. For a more detailed discussion of the relationship between these two objectives, see \Cref{appendix:vf_aux_true}.

\begin{figure*}
\centering
\includegraphics[width=\textwidth]{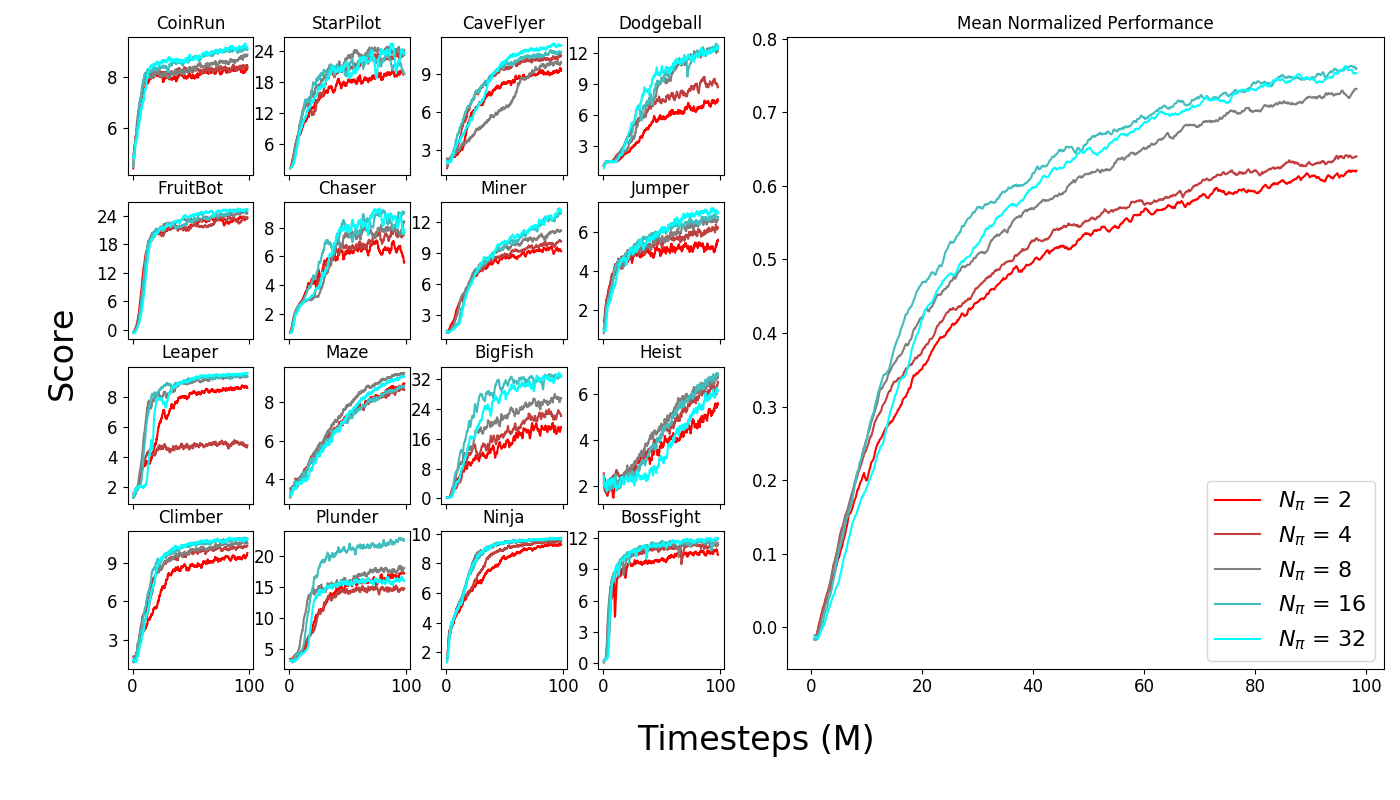}
\caption{Performance with varying auxiliary phase frequency}
\label{fig:npr}
\end{figure*}

\subsection{Auxiliary Phase Frequency}

We next investigate alternating between policy and auxiliary phases at different frequencies, controlled by the hyperparameter $N_{\pi}$. As described in \Cref{sec:algorithm}, we perform each auxiliary phase after every $N_{\pi}$ policy updates. We vary this hyperparameter from $2$ to $32$ and report results in \Cref{fig:npr}.

It is clear that performance suffers when we perform auxiliary phases too frequently. We conjecture that each auxiliary phase interferes with policy optimization, and that performing frequent auxiliary phases exacerbates this effect. It's possible that future research will uncover more clever optimization techniques to mitigate this interference. For now, we conclude that relatively infrequent auxiliary phases are critical to success.

\begin{figure*}
\centering
\includegraphics[width=\textwidth]{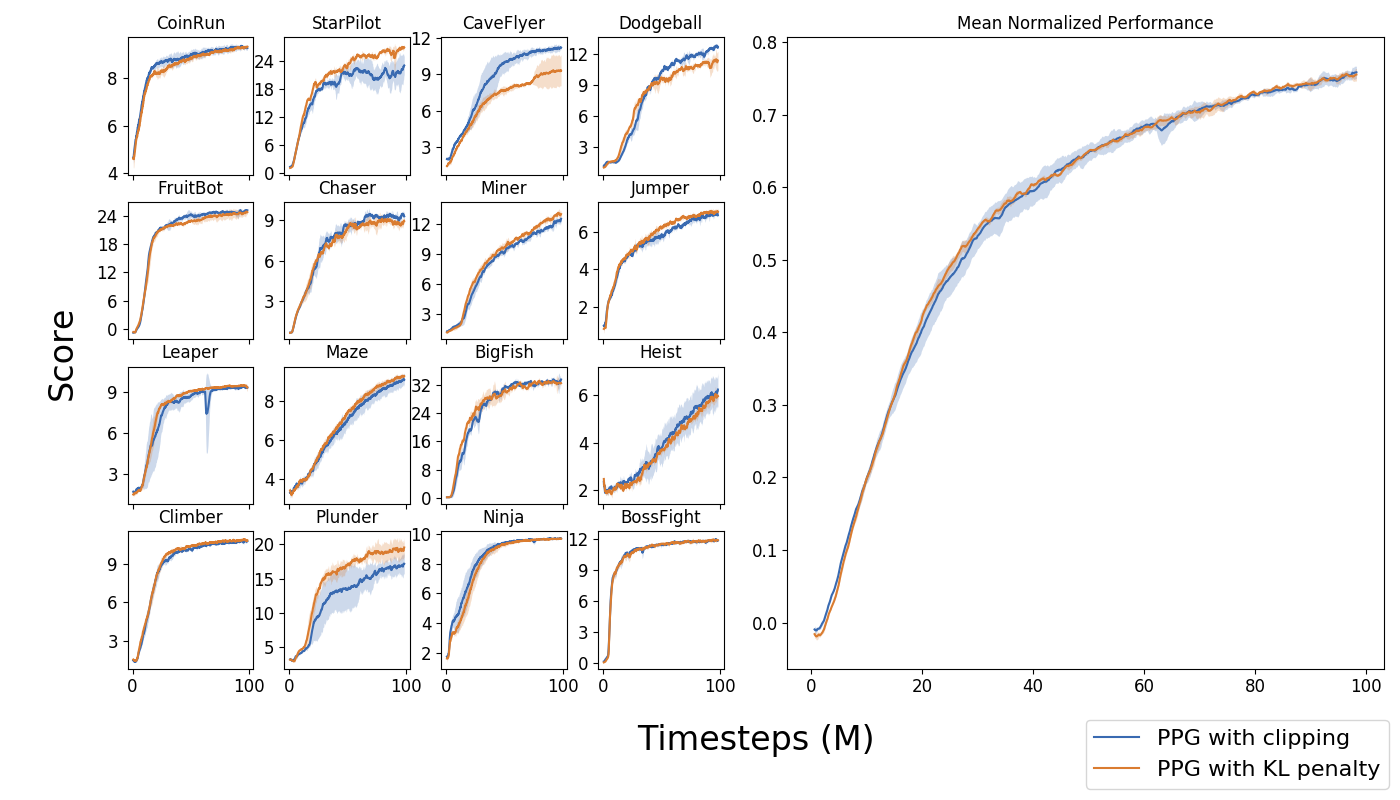}
\caption{The impact of replacing the clipping objective ($L^{clip}$) with a fixed KL penalty objective ($L^{KL}$)}
\label{fig:kl_penalty}
\end{figure*}

\subsection{KL Penalty vs Clipping}

As an alternative to clipping, \cite{ppo} proposed using an adaptively weighted KL penalty. We now investigate the use of a KL penalty in PPG, but we instead choose to keep the relative weight of this penalty fixed. Specifically, we set the policy gradient loss (excluding the entropy bonus) to be $$L^{KL} = \mathbb{\hat{E}}_t\left[-\hat{A_t}\frac{\pi_{\theta}(a_t|s_t)}{\pi_{\theta_{old}}(a_t|s_t)} + \beta_{\pi} \cdot  KL[\pi_{\theta_{old}}(\cdot|s_t),\pi_{\theta}(\cdot|s_t)]\right]$$ where $\beta_{\pi}$ controls the weight of the KL penalty. After performing a hyperparameter sweep, we set $\beta_{\pi}$ to $1$. Results are shown in \Cref{fig:kl_penalty}. We find that a fixed KL penalty objective performs remarkably similarly to clipping when using PPG. We suspect that using clipping (or an adaptive KL penalty) is more important when rewards are poorly scaled. We avoid this concern by normalizing rewards so that discounted returns have approximately unit variance. In any case, we highlight the effectiveness of the KL penalty variant of PPG since $L^{KL}$ is arguably easier to analyze than $L^{clip}$, and since future work may wish to build upon either objective.

\begin{figure*}
\centering
\includegraphics[width=\textwidth]{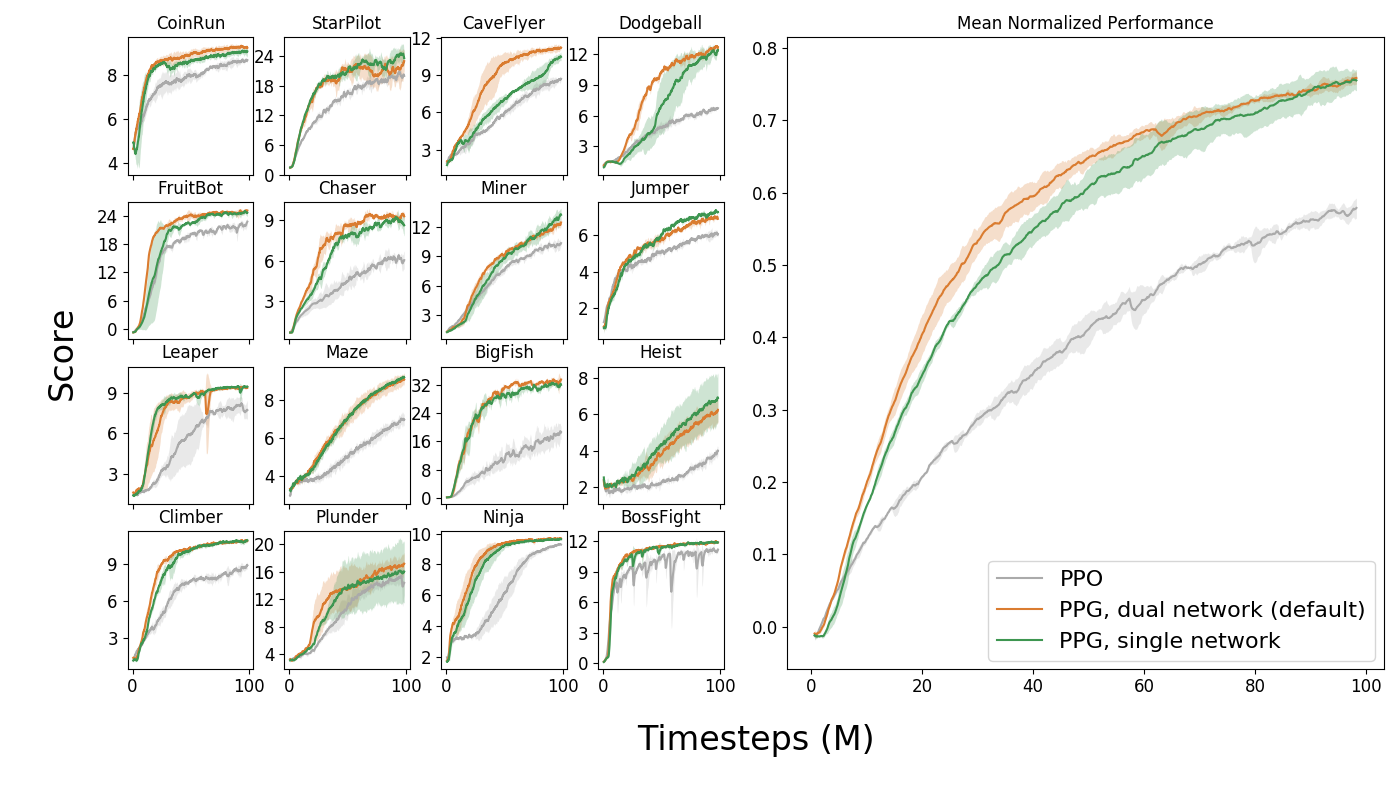}
\caption{A comparison between the default implementation of PPG which trains two separate networks, and a single-network variant that mimics the same training dynamics by detaching the gradient when necessary. PPO shown for reference.}
\label{fig:detach}
\end{figure*}

\subsection{Single-Network PPG}

By default, PPG comes with an increased memory footprint. Since we use disjoint policy and value function networks instead of a single unified network, we use approximately twice as many parameters compared to the PPO baseline. We can recover this cost and maintain most of the key benefits of PPG by using a single network that appropriately detaches the value function gradient. During the policy phase, we detach the value function gradient at the last layer shared between the policy and value heads, preventing the value function gradient from influencing shared parameters. During the auxiliary phase, we take the value function gradient with respect to all parameters, including shared parameters. This allows us to benefit from the representations learned by the value function, while still removing the interference during the policy phase.

As we can see, using PPG with this single shared network performs almost as well as PPG with a dual network architecture. We were initially concerned that the value function might be unable to train well during the policy phase with the detached gradient, but in practice this does not appear to be a major problem. We believe this is because the value function can still train from the full gradient during the auxiliary phase.

\section{Related Work}

\cite{igl2020impact} recently proposed Iterative Relearning (ITER) to reduce the impact of non-stationarity during RL training. ITER and PPG share a striking similarity: both algorithms alternate between a standard RL phase and a distillation phase. However, the nature and purpose of the distillation phase varies. In ITER, the policy and value function teachers are periodically distilled into newly initialized student networks, in an effort to improve generalization. In PPG, the value function network is periodically distilled into the policy network, in an effort to improve sample efficiency.

Prior work has considered the role the value function plays as an auxiliary task. \cite{bellemare2019geometric} investigate using value functions to train useful representations, specifically focusing on a special class of value functions called Adversarial Value Functions (AVFs). They find that AVFs provide a useful auxiliary objective in the four-room domain. \cite{lyle2019comparative} suggest that the benefits of distributional RL \citep{bellemare2017distributional} can perhaps be attributed to the rich signal the value function distribution provides as an auxiliary task. We find that the representation learning performed by the value function is indeed critical in Procgen environments, although we consider only the value function of the current policy, and we do not model the full value distribution.

Off-policy algorithms like Soft Actor-Critic (SAC) \citep{sac}, Deep Deterministic Policy Gradient (DDPG) \citep{ddpg}, and Actor-Critic with Experience Replay (ACER) \citep{acer} all employ replay buffers to improve sample efficiency via off-policy updates. PPG also utilizes a replay buffer, specifically when performing updates during the auxiliary phase. However, unlike these algorithms, PPG does not attempt to improve the policy from off-policy data. Rather, this replay buffer data is used only to better fit the value targets and to better train features for the policy. SAC also notably uses separate policy and value function networks, presumably, like PPG, to avoid interference between their respective objectives.

Although we use the clipped surrogate objective from PPO \citep{ppo} throughout this work, PPG is in principle compatible with the policy objectives from any actor-critic algorithm. \cite{andrychowicz2020matters} recently performed a rigorous empirical comparison of many relevant algorithms in the on-policy setting. In particular, AWR \citep{awr} and V-MPO \citep{vmpo} propose alternate policy objectives that move the current policy towards one which weights the likelihood of each action by the exponentiated advantage of that action. Such objectives could be used in PPG, in place of the PPO objective.

There are also several trust region methods, similar in spirit to PPO, that would be compatible with PPG. Trust Region Policy Optimization (TRPO) \citep{trpo} proposed performing policy updates by optimizing a surrogate objective, whose gradient is the policy gradient estimator, subject to a constraint on the KL-divergence between the original policy and the updated policy. Actor Critic using Kronecker-Factored Trust Region (ACKTR) \citep{acktr} uses Kronecker-factored approximated curvature (K-FAC) to perform a similar trust region update, but with a computational cost comparable to SGD. Both methods could be used in the PPG framework.

\section{Conclusion}

The results in \Cref{sec:policy_sr} and \Cref{sec:value_sr} make it clear that the optimal level of sample reuse varies significantly between the policy and the value function. Training these two objectives with varying sample reuse is not possible in a conventional actor-critic framework using a shared network architecture. By decoupling policy and value function training, PPG is able to reap the benefits of additional value function training without significantly interfering with the policy. To achieve this, PPG does introduce several new hyperparameters, which creates some additional complexity relative to previous algorithms. However, we consider this a relatively minor cost, and we note that the chosen hyperparameter values generalize well across all 16 Procgen environments.

By mitigating interference between the policy and the value function while still maintaining the benefits of shared representations, PPG significantly improves sample efficiency on the challenging Procgen Benchmark. Moreover, PPG establishes a framework for optimizing arbitrary auxiliary losses alongside RL training in a stable manner. We have focused on the value function error as the sole auxiliary loss in this work, but we consider it a compelling topic for future research to evaluate other auxiliary losses using PPG.

\changeurlcolor{black}

\bibliography{phasic_policy_gradient}

\changeurlcolor{blue}

\appendix

\clearpage

\section{Hyperparameters} \label{appendix:hyperparameters}

We use the Adam optimizer \citep{kingma2014adam} in all experiments.

\subsection{PPG-Specific Hyperparameters}

\begin{center}
 \begin{tabular}{||c c||} 
 \hline
 $N_{\pi}$ & 32 \\ 
 \hline
 $E_{\pi}$ & 1 \\
 \hline
 $E_V$ & 1 \\
 \hline
 $E_{aux}$ & 6 \\ 
 \hline
 $\beta_{clone}$ & 1 \\
 \hline
 \# minibatches per aux epoch per $N_{\pi}$ & 16 \\
 \hline
\end{tabular}
\end{center}

\subsection{Other Hyperparameters}

\begin{center}
 \begin{tabular}{||c c||} 
 \hline
 $\gamma$ & .999 \\ 
 \hline
 $\lambda$ & .95 \\
 \hline
 \# timesteps per rollout & 256 \\
 \hline
 \# minibatches per epoch & 8 \\ 
 \hline
 Entropy bonus coefficient ($\beta_S$) & .01 \\
 \hline
 PPO clip range ($\epsilon$) & .2 \\
 \hline
 Reward Normalization? & Yes \\
 \hline
 Learning rate & \num{5e-4} \\
 \hline
 \# workers & 4 \\
 \hline
 \# environments per worker & 64 \\ 
 \hline
 Total timesteps & 100M \\
 \hline
 LSTM? & No \\
 \hline
 Frame Stack? & No \\
 \hline
\end{tabular}
\end{center}

\newpage
\section{Shared vs Separate Networks} \label{appendix:separate_networks}

\begin{figure*}[h]
\centering
\includegraphics[width=\textwidth]{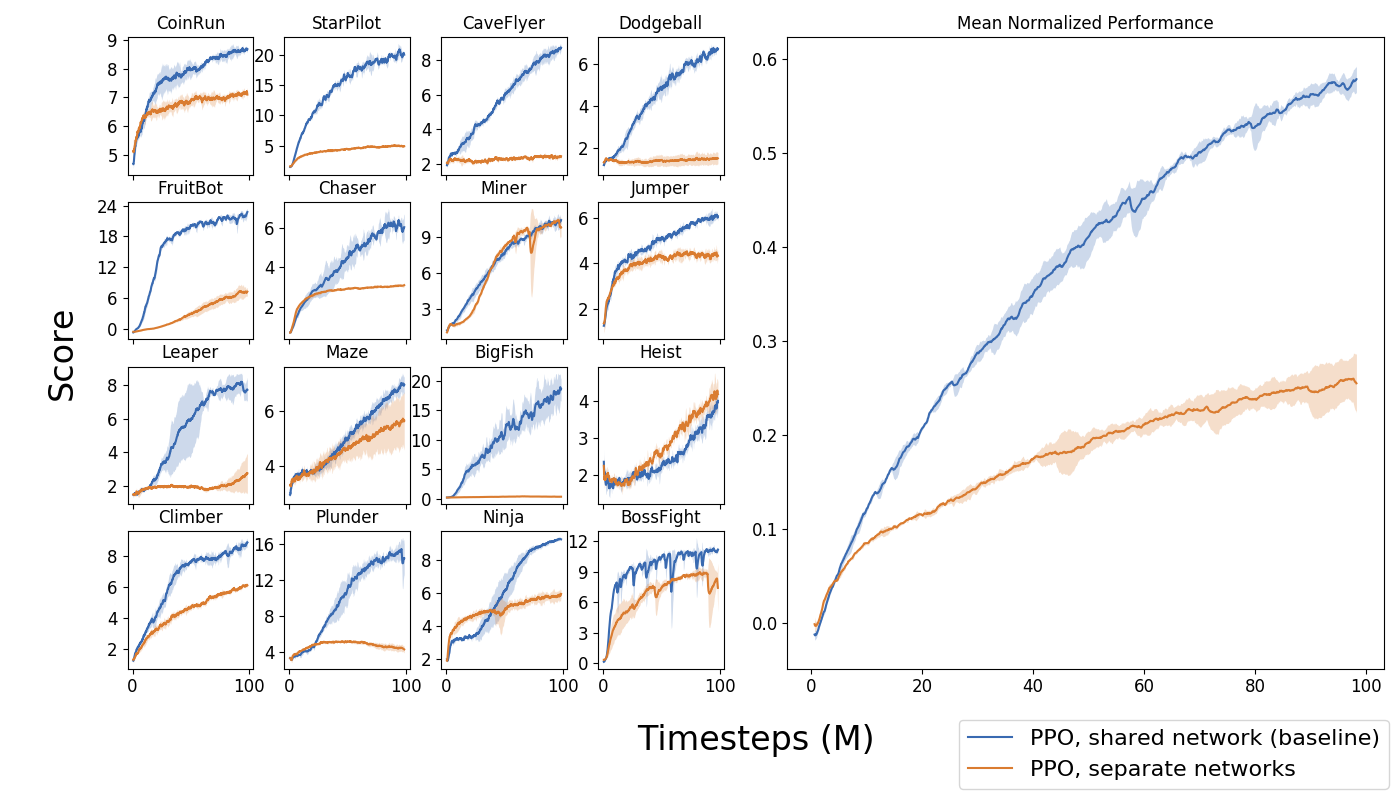}
\caption{A comparison between two implementations of PPO on Procgen Benchmark. The baseline shares features between the policy and value networks, while the ablation trains separate policy and value networks.}
\label{fig:pposep}
\end{figure*}

\newpage
\section{Auxiliary Phase Value Function Training} \label{appendix:vf_aux_true}

\begin{figure*}[h]
\centering
\includegraphics[width=\textwidth]{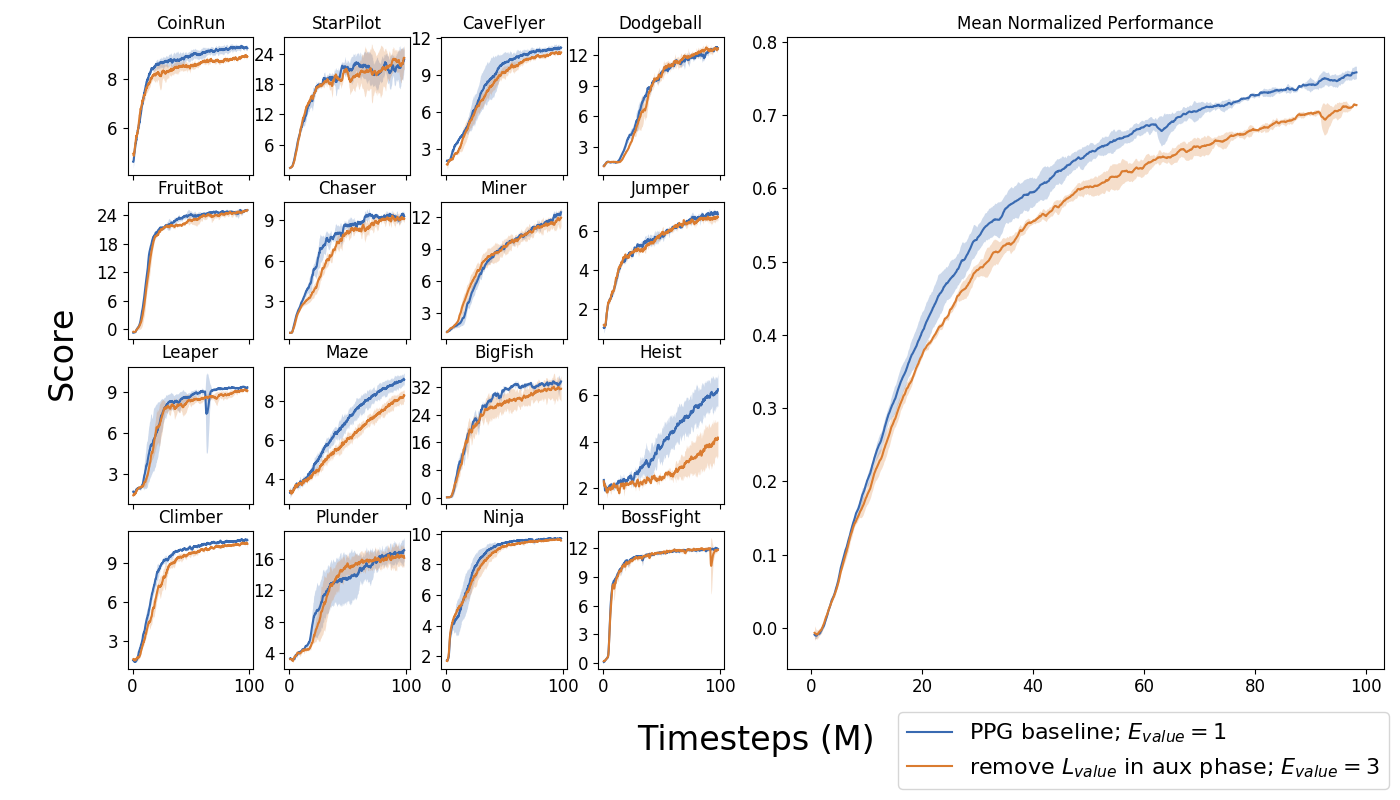}
\caption{The performance of a variant of PPG which skips the optimization of $L^{value}$ during the auxiliary phase, in favor of additional optimization of $L^{value}$ during the policy phase.}
\label{fig:ppgnoaux}
\end{figure*}

We now discuss the relative importance of optimizing $L^{value}$ and $L^{joint}$ during the auxiliary phase. From \Cref{appendix:separate_networks}, we know that $L^{joint}$ is crucial; without some optimization of this objective, there is no mechanism to share features between the value function and the policy. Although it is convenient to optimize $L^{value}$ during the auxiliary phase as well, it is not strictly necessary. It is also viable to perform extra value function optimization during the policy phase (by increasing $E_V$), while removing the optimization of $L^{value}$ from the auxiliary phase. A comparison between this variant and the PPG baseline are shown in \Cref{fig:ppgnoaux}. Although the PPG baseline has a slight advantage, we can see that the choice to optimize $L^{value}$ during the auxiliary phase is not an essential element of PPG.

\newpage
\section{PPO Sample Reuse} \label{appendix:ppo_sample_reuse}

\begin{figure*}[h]
\centering
\includegraphics[width=\textwidth]{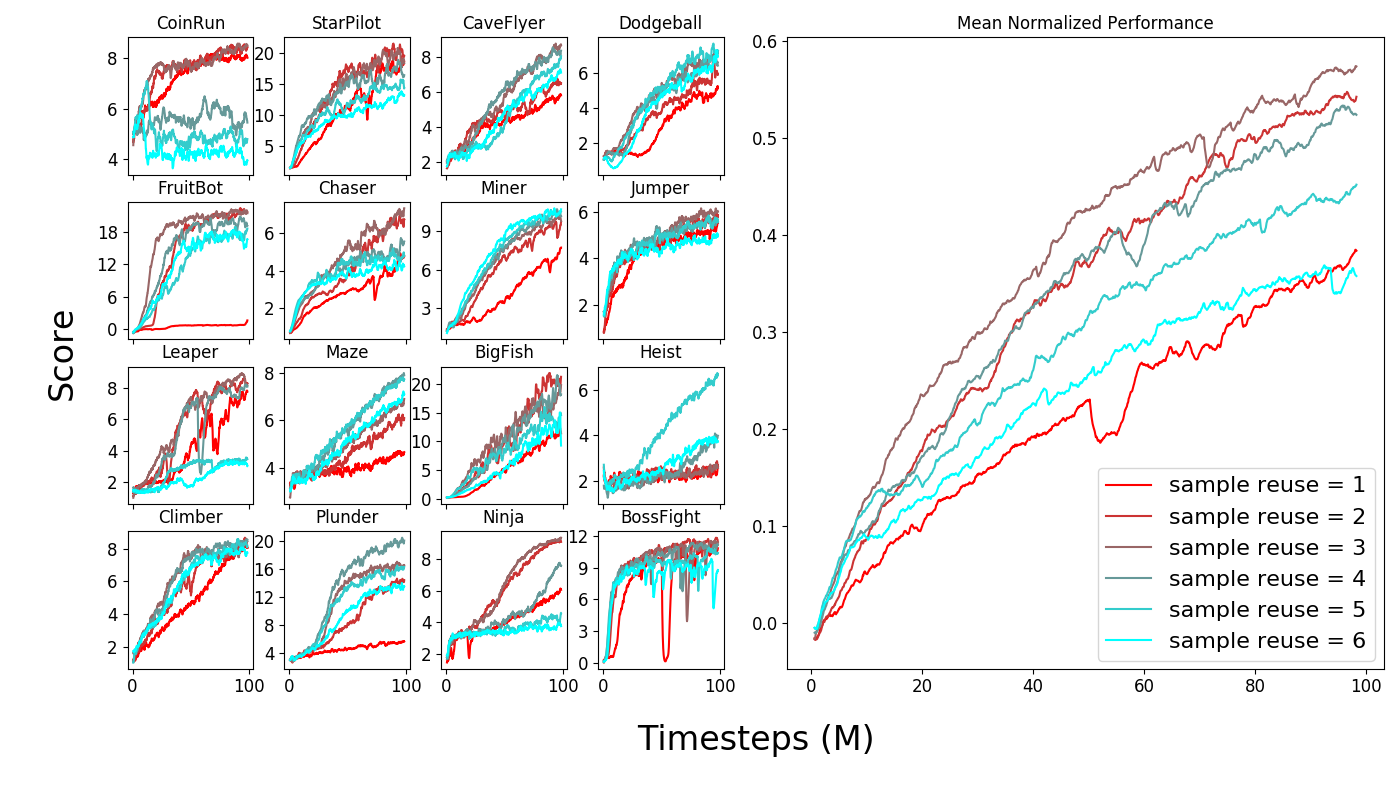}
\caption{A comparison between different levels of sample reuse in PPO.}
\label{fig:pposr}
\end{figure*}

We sweep over the different values for sample reuse in PPO, from 1 to 6. Empirically, we find that a sample reuse of 3 is optimal, given our other hyperparameter settings. As discussed in \Cref{sec:policy_sr}, the results with PPG suggest that the poor performance of PPO with low sample reuse is due to the fact that the value function, not the policy, is being under-trained.

\end{document}